\documentclass[conference]{IEEEtran}
\IEEEoverridecommandlockouts
\usepackage{cite}

\usepackage{amsmath,amssymb,amsfonts}
\usepackage{algorithmic}
\usepackage{graphicx}
\usepackage{textcomp}
\usepackage{tabularx}
\usepackage{multirow} 
\usepackage{multicol}
\usepackage{xcolor}
\usepackage{multirow}
\usepackage{caption}
\usepackage{algorithm}
\usepackage{algorithmic}
\usepackage{siunitx}
\usepackage{array}
\usepackage{multirow}
\def\BibTeX{{\rm B\kern-.05em{\sc i\kern-.025em b}\kern-.08em
    T\kern-.1667em\lower.7ex\hbox{E}\kern-.125emX}}
\begin{document}

\title{Scalability Matters: Overcoming Challenges in InstructGLM with Similarity-Degree-Based Sampling\\
}

\author{%
Hyun Lee\IEEEauthorrefmark{1}\IEEEauthorrefmark{2}\thanks{\IEEEauthorrefmark{1}These authors contributed equally to this work.\protect\\\IEEEauthorrefmark{2}Corresponding author: hyun.lee@trincoll.edu}, 
Chris Yi\IEEEauthorrefmark{1}, 
B.D.S. Aritra, 
Maminur Islam \\
Dept. of Computer Science, Trinity College, Hartford, Connecticut, USA \\
\{hyun.lee, christopher.yi, bds.aritra, maminur.islam\}@trincoll.edu
}

\maketitle

\begin{abstract}

Large Language Models (LLMs) have demonstrated strong capabilities in various natural language processing tasks; however, their application to graph-related problems remains limited, primarily due to scalability constraints and the absence of dedicated mechanisms for processing graph structures. Existing approaches predominantly integrate LLMs with Graph Neural Networks (GNNs), using GNNs as feature encoders or auxiliary components. However, directly encoding graph structures within LLMs has been underexplored, particularly in the context of large-scale graphs where token limitations hinder effective representation.
To address these challenges, we propose SDM-InstructGLM, a novel instruction-tuned Graph Language Model (InstructGLM) framework that enhances scalability and efficiency without relying on GNNs. Our method introduces a similarity-degree-based biased random walk mechanism, which selectively samples and encodes graph information based on node-feature similarity and degree centrality, ensuring an adaptive and structured representation within the LLM. This approach significantly improves token efficiency, mitigates information loss due to random sampling, and enhances performance on graph-based tasks such as node classification and link prediction. Furthermore, our results demonstrate the feasibility of LLM-only graph processing, enabling scalable and interpretable Graph Language Models (GLMs) optimized through instruction-based fine-tuning. This work paves the way for GNN-free approaches to graph learning, leveraging LLMs as standalone graph reasoning models. Our source code is available on GitHub\footnote{https://github.com/mhsrn21/SDM-InstructGLM}.
\end{abstract}


\section{Introduction}

LLMs (Large Language Models) \cite{BERT,LLaMA} have shown remarkable advancements in understanding and generating natural language.While most existing models are primarily designed to process sequential text, recent developments have expanded their capabilities to handle multi-modal data\cite{RadFord,AlaY,LI,YU}. Among these advancements, there has been increasing interest in applying LLMs to process graph-structured data. For instance, Graph-Bert \cite{zhang2020graphbertattentionneededlearning} and GraphFormers \cite{ying2021transformersreallyperformbad} have demonstrated the potential of transformer-based models for graph representation learning, while recent works\cite{sun2025graphiclunlockinggraphlearning,xu2024llmgnncomplementarydistilling,Liu_2024,li2024surveygraphmeetslarge,LEE} have explored the integration of LLMs with graph data for tasks such as node classification, link prediction., graph reasoning, and molecular graph analysis.

However, despite these advancements, the field of Graph Representation Learning using LLMs still faces several critical challenges that have spurred ongoing research to address their limitations. LLMs currently struggle to effectively encode and interpret the structural information inherent in graphs\cite{ying2021transformersreallyperformbad,zhang2020graphbertattentionneededlearning}. This includes understanding the relationships and interconnections between graph nodes, which poses a significant obstacle to fully utilizing LLMs in graph-based tasks.

As a result, most existing works focus on addressing these challenges by pairing LLMs with auxiliary models to solve graph-related problems more effectively. Among these, GNNs (Graph Neural Networks) have emerged as the most prominent auxiliary model, and much of the current research is centered on integrating GNNs with LLMs\cite{author2023gnnllm,chen2024LLaGA}. However, these GNN-LLM architectures come with inherent limitations that hinder their performance and scalability, such as the issue of over-smoothing in GNNs \cite{cai2020oversmoothing,oversmoothing}.

To address these challenges and further narrow the gap between graphs and natural languages, models such as InstructGLM \cite{InstructGLM} have been proposed, allowing generative graph learning by learning on graph structures purely using natural languages. However, limitations on input token lengths inherent to LLMs make it difficult for such approaches to learn large graphs, and measures of overcoming such problems have been relatively understudied.

To tackle these issues, we introduce a similarity-degree-based biased random walk mechanism (SDM-InstructGLM), which selectively explores highly relevant nodes based on node-feature similarity and degree centrality. This approach efficiently encodes local and global graph structures within LLMs while mitigating the scalability issues imposed by token length constraints.

\begin{figure}[t]
    \centering
    \includegraphics[width=0.45\textwidth]{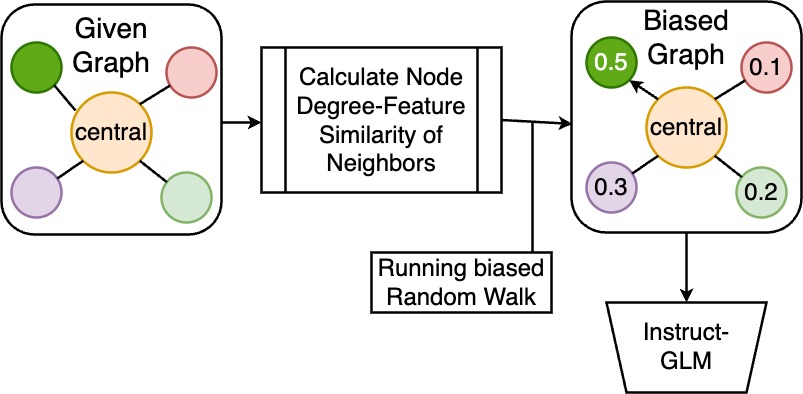}
    \caption{Illustration of the SDM-InstructGLM framework, which enhances LLM training on graph-structured data using a similarity-degree-based random walk to preserve key structural relationships.}
    \label{fig:SDM_InstructGLM}
\end{figure}

As depicted in Figure \ref{fig:SDM_InstructGLM}, our mechanism consists of three main stages:

\begin{enumerate}
    \item \textbf{Node Selection via Similarity and Degree Centrality}:  
    We identify key nodes by computing a cumulative similarity-degree score, ensuring the retention of both structurally significant and semantically relevant nodes.

    \item \textbf{Biased Random Walk Mechanism}:  
    Our method prioritizes exploration of graph regions with high structural importance, biasing transitions toward nodes with stronger feature similarity and higher degree centrality. This enhances both local and global graph connectivity.

    \item \textbf{Token-Efficient Graph Encoding}:  
    Selected nodes and their relationships are transformed into a structured textual representation optimized for LLM processing, preserving essential graph features while minimizing token redundancy.
\end{enumerate}

\section{Related Work}

Graphs serve as fundamental data structures for representing relationships in various domains, including social networks, molecular biology, and recommendation systems. Traditional Graph Neural Network (GNN) models such as GCN or GAT \cite{GCN, GAT} have demonstrated strong performance in tasks such as node classification and link prediction by leveraging message-passing mechanisms. However, they suffer from several limitations, including scalability issues\cite{GraphSage}, over-smoothing\cite{cai2020oversmoothing,oversmoothing} in deep architectures, and limited generalization across unseen graphs. To address computational constraints, GraphSAGE \cite{GraphSage} introduced random neighbor sampling, allowing models to aggregate information from a subset of neighbors instead of the full graph. While this approach improves efficiency, it also results in information loss by neglecting crucial topological structures.

Recent advancements in Large Language Models (LLMs) have opened new possibilities for modeling structured data, including graphs. Models such as GPT-4 \cite{GPT4}, Flan-T5 \cite{Flan}, and LLaMA \cite{LLaMA} exhibit strong generalization capabilities across diverse tasks. Researchers have begun exploring LLMs for graph learning by transforming graph structures into natural language prompts \cite{gpt4graph, llmgraphsurvey}. This reformulation enables LLMs to learn graph representations without explicit message passing. However, a major limitation of LLM-based approaches is their token length constraint, which restricts the amount of structural information that can be encoded per input. This limits their ability to model large-scale graphs effectively.

To bridge the gap between GNNs and LLMs, hybrid GNN-LLM frameworks have emerged, integrating the strengths of both paradigms. These methods can be broadly categorized into the following approaches:

\begin{itemize}
    \item \textbf{GNNs as Prefix:} GNNs preprocess graph data and convert it into structured tokens that are then fed into LLMs for inference. This includes node-level tokenization (e.g., GraphGPT \cite{GraphGPT}, HiGPT \cite{HiGPT}) and graph-level tokenization (e.g., GraphLLM \cite{GraphLLM}).
    \item \textbf{LLMs as Prefix:} LLMs generate embeddings or labels that serve as inputs to GNNs. For instance, G-Prompt \cite{GPrompt} uses LLM-derived embeddings for downstream GNN-based tasks, while OpenGraph \cite{xia2024opengraph} employs LLMs to generate synthetic graph labels.
    \item \textbf{LLMs-Graphs Integration:} Some methods aim for deeper integration between LLMs and GNNs via joint training or contrastive learning. Notable examples include GreaseLM \cite{zhang2022greaselm} and DGTL \cite{qin2024dgtl}, which fuse transformer layers with GNN components.
    \item \textbf{LLMs-Only:} InstructGLM \cite{InstructGLM}, WalkLM \cite{tan2023walklm,ourlastpaper}, and GraphWiz \cite{chen2024graphwiz} explore training LLMs directly on graph data, eliminating the need for specialized graph neural networks.
\end{itemize}

While these methods show promise, they still face scalability issues. LLMs have a fixed token length, making it difficult to represent large-scale graphs beyond a few hops. Hybrid approaches often inherit the computational overhead of both GNNs and LLMs, leading to inefficiencies in real-world deployment. Moreover, prompt engineering remains an open challenge, as encoding structural information into text can introduce ambiguity.

\section{Preliminary}

\subsection{Graph Representation}

A graph is formally represented as:
\begin{equation}
    G = (V, A, E, \{ N_v \}_{v \in V}, \{ E_e \}_{e \in E})
\end{equation}
where \(V\) is the set of nodes, \(E \subseteq V \times V\) is the set of edges, \(A \in \{0,1\}^{|V| \times |V|}\) is the adjacency matrix, and \(N_v\) and \(E_e\) represent node and edge features, respectively.  

The \(k\)-hop neighborhood of a node \( v \) is defined as:
\begin{equation}
    A_v^k = \{ u \in V \mid d(v, u) \leq k \}
\end{equation}

\subsection{Graph Neural Networks}

Graph Neural Networks (GNNs) have been widely adopted for node classification due to their ability to iteratively aggregate local neighborhood information while preserving the graph’s structural relationships. Among different GNN architectures, Graph Attention Networks (GATs) and Graph Transformers introduce attention-based aggregation to enhance the learning of important node interactions.

GATs \cite{GAT} incorporate an attention mechanism that assigns adaptive weights to neighboring nodes, enabling selective aggregation during message passing. The attention coefficient \(\alpha_{ij}\) for a target node \(i\) and its neighbor \(j\) is computed as:

\begin{equation}
    \alpha_{ij} = \frac{\exp(\text{LeakyReLU}(\mathbf{a}^T [\mathbf{h}_i || \mathbf{h}_j]))}{\sum_{k \in \mathcal{N}(i)} \exp(\text{LeakyReLU}(\mathbf{a}^T [\mathbf{h}_i || \mathbf{h}_k]))}
\end{equation}

where \(\mathbf{h}_i\) represents the feature embedding of node \(i\), \(\mathcal{N}(i)\) denotes its neighborhood, and \(\mathbf{a}\) is a trainable attention vector. This mechanism enhances node representations by assigning higher importance to informative neighbors.

Graph Transformers \cite{dwivedi2021generalization,ying2021transformers} extend this approach by leveraging global self-attention, enabling information exchange beyond local neighborhoods. Unlike message-passing GNNs, which depend on adjacency structures, Graph Transformers compute attention weights across all node pairs:

\begin{equation}
    \mathbf{h}_i' = \sum_{j \in V} \text{softmax} \left( \frac{\mathbf{h}_i \mathbf{W}_Q (\mathbf{h}_j \mathbf{W}_K)^T}{\sqrt{d}} \right) \mathbf{h}_j \mathbf{W}_V
\end{equation}

Here, \(\mathbf{W}_Q, \mathbf{W}_K, \mathbf{W}_V\) are trainable weight matrices, and \(d\) is the embedding dimension. This mechanism enables the model to effectively capture both local and long-range dependencies in graph structures.

\subsection{Instruction-Based Graph Representation}

Unlike GNNs, InstructGLM encodes graph topology as structured natural language, converting node and edge relationships into textual prompts. Given a graph \( G = (V, E) \), a textual representation \( I \) is constructed as:

\begin{equation}
    I = T(v, A, \{N_v\}_{v \in V}, \{E_e\}_{e \in E})
\end{equation}

where \( T(\cdot) \) is a function that transforms graph structure into a descriptive format. This enables LLMs to process graph data without explicit adjacency-based computations.

A 2-hop neighborhood representation for node \( v \) is expressed as:

\begin{equation}
    T(v, A) = \text{``$v$ connects with $\{ v_2 \mid v_2 \in A_v^2 \}$ in 2 hops"}
\end{equation}

where \( A_v^2 \) denotes the set of nodes reachable from \( v \) within two hops. If node and edge attributes are included, the textual description is extended as:

\small
\begin{equation}
    \text{\scriptsize}
    T(v, A, N_v, E_e) = \text{``($v, N_v$) links to $\{ (v_2, N_{v_2}) \}$ via $\{ (v_1, N_{v_1}) \}$"}
    \normalsize
\end{equation}

While this structured approach allows LLMs to interpret graph structures, it remains inherently constrained by token length limitations.

\subsection{Generative Instruction Tuning for Node Classification}

InstructGLM formulates node classification as a generative process, where the model predicts class labels autoregressively. Given an input query \( x \) and previously predicted labels \( y_{<j} \), the conditional probability is defined as:

\begin{equation}
    P_\theta (y_j \mid x, y_{<j}) = \text{LLM}_\theta(x, y_{<j})
\end{equation}

where the input sequence \( x \) consists of:
\begin{itemize}
    \item \( P \): the instruction prefix,
    \item \( I \): the graph description prompt, and
    \item \( Q \): the classification query (e.g., ``Which category should \( v \) be classified as?").
\end{itemize}

Training is performed using the Negative Log-Likelihood (NLL) loss:

\begin{equation}
    L_\theta = -\sum_{j=1}^{|y|} \log P_\theta (y_j \mid x, y_{<j})
\end{equation}

This approach enables the model to generate class labels conditioned on structured graph descriptions.

\subsection{InstructGLM Limitations}

Despite its effectiveness in leveraging natural language representations for graph processing, InstructGLM faces significant scalability challenges due to the inherent token length constraints of LLMs. Models such as Flan-T5 (512 tokens) and LLaMA (2048 tokens) impose strict input length restrictions, limiting the amount of structural information that can be encoded within a single prompt. This constraint significantly impacts the ability to represent large-scale graphs, where capturing multi-hop relationships and extensive connectivity patterns is essential.

To tackle these constraints, InstructGLM adopts a random neighbor sampling strategy, similar to GraphSAGE \cite{GraphSage}, where a fixed number of neighbors is selected rather than encoding the entire graph. While this method enables InstructGLM to process larger graphs, it introduces information loss due to arbitrary neighbor selection. Since neighbors are sampled randomly, the model may exclude structurally significant nodes, disrupting the graph’s topological coherence and omitting key relationships.



\section{SDM-InstructGLM}

This section presents the proposed SDM-InstructGLM framework, which enhances the ability of large language models (LLMs) to process graph-structured data. The framework consists of three primary components: (1) a biased random walk mechanism that incorporates node-feature similarity and structural importance, (2) a structured node ordering strategy, and (3) a hop-aware node selection method for efficient token utilization. The proposed approach eliminates the need for explicit attention mechanisms while ensuring that the sampled subgraphs retain both local and global contextual information.

\subsection{Biased Random Walk with Cosine Similarity and Degree}
Instead of utilizing a conventional attention mechanism, the proposed method mimics attention by performing a biased random walk. In this approach, the probability of transitioning to a neighboring node is influenced by both the cosine similarity between node embeddings and the degree centrality of the candidate node. Given a current node \( u \), the transition probability to a neighboring node \( v \) is computed as:

\begin{align}
z_{uv} &= \text{cosSim}(\mathbf{h}_u, \mathbf{h}_v) \cdot \mathrm{deg}(v) \\
&= \frac{\mathbf{h}_u \cdot \mathbf{h}_v}{\|\mathbf{h}_u\| \|\mathbf{h}_v\|} \cdot \mathrm{deg}(v)
\end{align}

\begin{align}
p(v \mid u) &= \mathrm{softmax}(z_{uv}) \\
&= \frac{e^{z_{uv}}}{\sum_{w \in \mathcal{N}(u)} e^{z_{uw}}}
\end{align}

\text{where:}
\begin{itemize}
    \item $\mathcal{N}(u)$ \text{denotes the set of neighbors of node } $u$
    \item $\text{cosSim}(\mathbf{h}_u, \mathbf{h}_v) = \frac{\mathbf{h}_u \cdot \mathbf{h}_v}{\|\mathbf{h}_u\| \|\mathbf{h}_v\|}$ \text{ represents the cosine similarity}\\
    \text{between feature embeddings } $\mathbf{h}_u$ \text{ and } $\mathbf{h}_v$
    \item $\mathrm{deg}(v)$ \text{ denotes the degree of node } $v$\text{, which serves as}\\
    \text{a measure of its structural importance.}
\end{itemize}

\begin{figure}[t]
    \centering
    \includegraphics[width=0.45\textwidth]{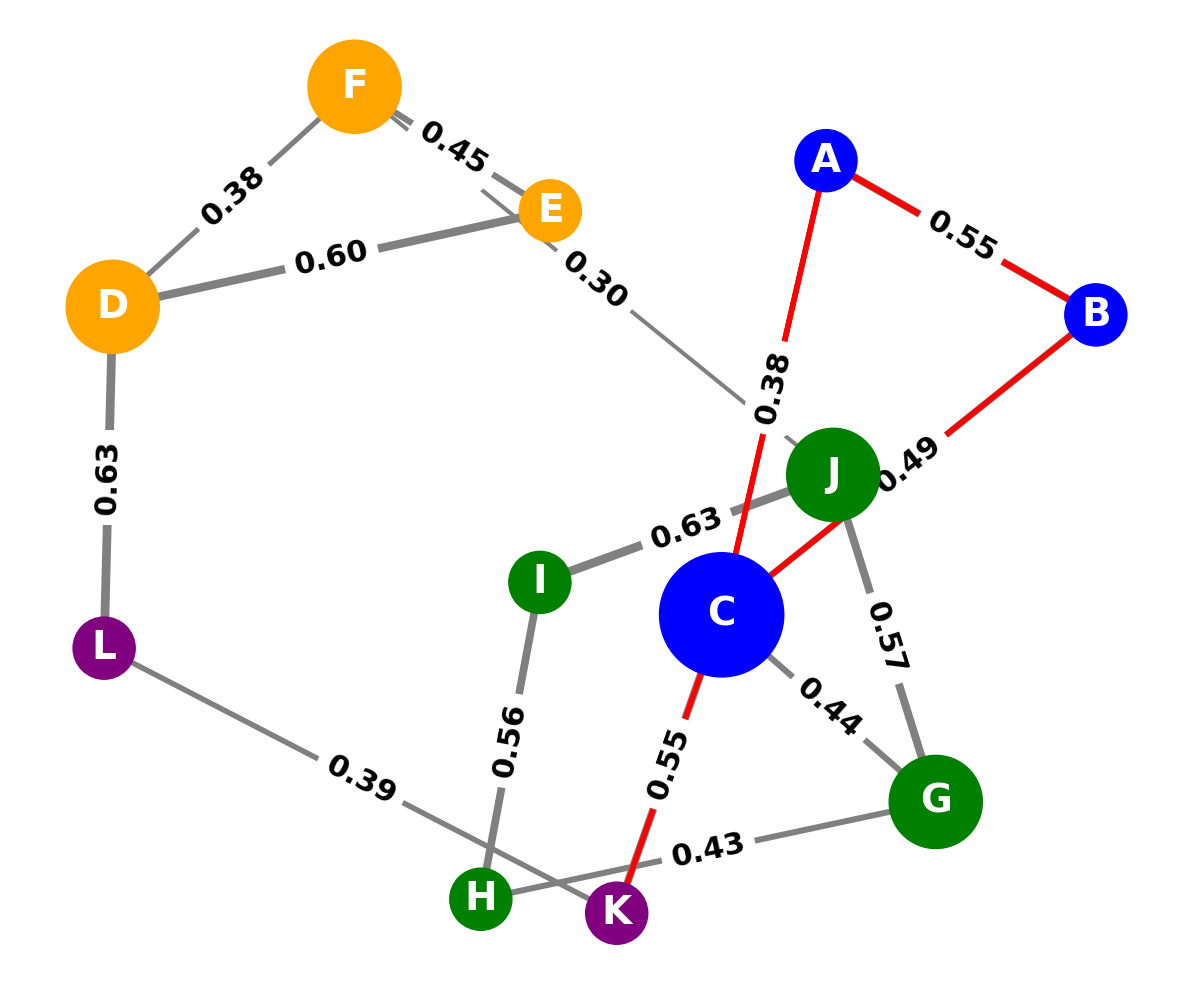}
    \caption{Illustration of a degree and similarity-biased random walk.}
    \label{fig:Biased_RW}
\end{figure}

Figure~\ref{fig:Biased_RW} illustrates the biased random walk process, where the transition probabilities are influenced by both node degree and cosine similarity. Nodes are color-coded based on their respective categories, with node sizes representing their degree centrality. Edge thickness reflects cosine similarity, with thicker edges indicating stronger feature similarity. The red-highlighted path represents an example biased random walk starting from node A, where the probability of transitioning to a neighboring node is determined by both similarity and structural influence. This probability distribution ensures that nodes with both high feature similarity and greater structural importance are more likely to be selected during the walk. The approach provides a principled mechanism for exploring graph neighborhoods, effectively preserving essential connectivity patterns while maintaining computational efficiency.

\subsection{Structured Node Ordering for Sequence Representation}

To enhance the sequence representation of graph data for LLM processing, the proposed method adopts a structured node ordering strategy. Instead of randomly ordering sampled nodes, the sequence is determined based on a combined measure of similarity and structural importance. Given a set of sampled nodes \( V' \), each node \( v \) is assigned a ranking score:

\[
s(v) = \text{cosSim}(\mathbf{h}_u, \mathbf{h}_v) \cdot \deg(v)
\]

where \( u \) is the anchor node, \( \mathbf{h}_u \) and \( \mathbf{h}_v \) are their respective feature embeddings, and \( \deg(v) \) represents the degree centrality of node \( v \). Nodes in \( V' \) are then sorted in descending order of \( s(v) \), ensuring that structurally and semantically significant nodes appear earlier in the sequence.

This structured ordering refines the input representation by prioritizing nodes that exhibit both high similarity to the anchor node and strong connectivity within the graph. Empirical results demonstrate that this method outperforms random ordering, as it allows the LLM to capture meaningful topological dependencies more effectively, leading to improved downstream task performance. This approach is also in line with earlier graph sampling techniques such as PinSage \cite{PinSage}.

\subsection{Hop-Aware Node Selection Under Token Constraints}

LLMs impose strict token length constraints, limiting the number of nodes that can be included in a sequence. To efficiently utilize the available token budget while maintaining a structurally representative subgraph, we introduce a hop-aware node sampling strategy. Instead of selecting a fixed number of nodes per hop or applying arbitrary truncation, the proposed method dynamically adjusts the allocation based on the empirical average node count per hop observed in the dataset. This ensures that the number of sampled nodes remains proportional to the graph’s inherent structure while preventing excessive token consumption.

Given a maximum token budget \( T_{\max} \), the number of nodes sampled at each hop level \( h \) is determined by:

\[
N_{\text{sample}}(h) = \min\left( N_{\text{hop}}(h), \frac{T_{\max}}{T_{\text{avg}}} \right)
\]

where:
\begin{itemize}
    \item \( N_{\text{hop}}(h) \) is the observed average number of nodes at hop distance \( h \) from the anchor node,
    \item \( T_{\text{avg}} \) is the estimated token cost per node, and
    \item \( T_{\max} \) is the total token budget allocated for node representations.
\end{itemize}

By dynamically capping the number of nodes sampled at each hop, this method ensures that the total token consumption remains within the allowed budget while preserving a balanced distribution of nodes across different hop distances. Unlike fixed-threshold sampling methods that impose rigid cutoffs, the hop-aware approach adapts to varying graph sizes by allocating nodes based on empirical hop distributions.

Empirical evaluations confirm that this structured sampling approach improves subgraph representation for LLMs by maintaining a stable token allocation strategy. By enforcing hop-wise node limits while avoiding arbitrary truncation, the method enables consistent and scalable sequence encoding regardless of graph size.

\section{Experiment}

To evaluate the effectiveness of SDM-InstructGLM, we conduct extensive experiments on two public datasets. Specifically, we aim to answer the following research questions:

\begin{itemize}
    \item \textbf{RQ1:} Compared to existing LLM-based graph representation methods, how does SDM-InstructGLM perform in preserving structural and semantic information?
    \item \textbf{RQ2:} How effective is each component of SDM-InstructGLM, including biased random walks, structured node ordering, and hop-aware node selection?
\end{itemize}

\subsection{Experiment Setup}

\begin{table}[t]
	\centering
	\caption{Dataset Statistics}
	\renewcommand\arraystretch{1.3}
	\scalebox{0.95}{%
	\begin{tabular}{c|c|c|c}
		\hline 
		\hline	
		\multicolumn{2}{c|}{\multirow{2}{*}{}} & \multicolumn{2}{c}{Dataset Names}\\ 
        \cline{3-4}
        \multicolumn{2}{c|}{} & \textsc{Cora} & \textsc{PubMed} \\
		\hline		
		\multicolumn{2}{c|}{\# Nodes} & 2708 & 19717 \\
        \multicolumn{2}{c|}{\# Edges} & 5429 & 44338 \\
        \multicolumn{2}{c|}{\# Classes} & 7 & 3 \\
        \multicolumn{2}{c|}{Diameter} & 19 & 18 \\
        \hline
        \multirow{2}{*}{\shortstack{\# 1-Hop \\ Neighbors}} & $\overline{x}$ & 3.90 & 4.50 \\
        ~ & $\sigma$ & 5.23 & 7.43 \\
        \hline
        \multirow{2}{*}{\shortstack{\# 2-Hop \\ Structures}} & $\overline{x}$ & 38.63 & 70.93 \\
        ~ & $\sigma$ & 53.70 & 130.29 \\
        \hline
        \multirow{2}{*}{\shortstack{\# 3-Hop \\ Structures}} & $\overline{x}$ & 279.66 & 1111.86 \\
        ~ & $\sigma$ & 421.77 & 2698.22 \\
		\hline
	\end{tabular}%
	}
	\label{dataset_stats}
\end{table}

\textbf{Dataset Description:}  To evaluate the effectiveness of our SDM-InstructGLM framework, we use two of the widely-used benchmark graph datasets, \textsc{Cora} and \textsc{PubMed} datasets \cite{yang2016}. Following the original InstructGLM experimental setup \cite{InstructGLM}, we use the raw text proposed in \cite{he2024}, as well as the same 60\%/20\%/20\% train/validation/test split. The statistics of the two datasets are summarized in Table \ref{dataset_stats}; we additionally provide the mean($\overline{x}$) and the standard deviation($\sigma$) of the number of 1-hop neighbors and 2-hop and 3-hop structures to demonstrate the relevance of the aforementioned token limit problem. We are unable to run the ArXiv dataset \cite{ogb} used in the original InstructGLM implementation due to issues with replicating the original code, a problem encountered by multiple users. Additionally, limitations with resources made it impractical to test on such a large dataset.

\textbf{Baseline:} We compare SDM-InstructGLM with its original backbone, InstructGLM, on the node classification task. While the original baseline was typically trained for 4 epochs, due to computational limitations, we conducted the experiment using only 2 epochs. 

\textbf{Evaluation Metric:} We assess the accuracy metric for the tests we conduct. Results are reported for SDM-InstructGLM using instruct fine-tuned \textit{Llama-7B}. For comparisons with Graph Neural Networks (GNNs) and Graph Transformers, refer to prior works on InstructGLM. Our findings indicate that SDM-InstructGLM consistently surpasses previous benchmarks, demonstrating its effectiveness as a foundation model for graph learning.

\textbf{Parameter Setting:} We employ a multi-prompt instruction-tuning framework for all experiments and report test accuracy as our primary metric. Instead of TF-IDF used by original \textsc{Cora} and \textsc{PubMed} datasets, which are insufficient to capture the semantic relationship between words, we employ BERT-based \cite{BERT} bidirectional embeddings to encode node features derived from textualized metadata (e.g., abstracts and titles). Unlike static embeddings (e.g., Word2Vec or GloVe), BERT dynamically captures contextual semantics by jointly attending to both preceding and following tokens. This is particularly beneficial for academic datasets such as \textsc{Cora} and \textsc{PubMed}, where domain-specific language and polysemy are prevalent. The bidirectional attention mechanism ensures high-fidelity semantic representation, leading to improved generalization when integrated with graph-based models. All experiments are conducted on a setup with eight A100 GPUs with 40GB VRAM each. Training takes about 3.0 hours per epoch for \textsc{PubMed}, and about 1.5 hours per epoch for \textsc{Cora}.

We set the learning rate to \num{8e-5} and the batch size to 4. We use the AdamW optimizer \cite{loshchilov2019} with a weight decay of 0. All experiments are conducted for 2 epochs.

\subsection{Performance Comparison (RQ1)}

To evaluate the ability of SDM-InstructGLM to preserve both structural and semantic information, we compare its performance with the original InstructGLM on node classification tasks across different hop distances. Table~\ref{accresult} reports accuracy results on the \textsc{Cora} and \textsc{PubMed} datasets when considering \textbf{1-hop, 2-hop, and 3-hop} structures.

\begin{table}[t]
	\centering
	\caption{ Accuracy Results
    }
	\renewcommand\arraystretch{1.3}
	\begin{tabular}{c|c|ccc}
		\hline 
		\hline	
		\multirow{2}*{Dataset} & \multirow{2}*{Model} & \multicolumn{3}{c}{Accuracy(\%) by \# Hops}\\ 
        \cline{3-5}
        ~ & ~ & 1 & 2 & 3\\
		\hline		
		\multirow{2}*{\textsc{Cora}}& InstructGLM (Original)  &  62.54 &  68.59 & 70.71 \\
		~ & SDM-InstructGLM   & \textbf{81.74}  & \textbf{84.92} & \textbf{83.81} \\
        \hline
        \multirow{2}*{\textsc{PubMed}}& InstructGLM (Original)& 91.56   & 90.87  & 90.85 \\
        ~ & SDM-InstructGLM   & \textbf{91.70}  & \textbf{91.48}   & \textbf{91.23} \\
		\hline
	\end{tabular}
		
	\label{accresult}
\end{table}

Across all settings, SDM-InstructGLM consistently outperforms InstructGLM, demonstrating its effectiveness in capturing both local and global context.

\subsubsection{Accuracy Improvements}
\noindent\textbf{\\\textsc{Cora} Dataset:} 
\begin{itemize}
    \item \textbf{1-hop:} SDM-InstructGLM achieves 81.74\%, outperforming InstructGLM by \textbf{19.2\%}.  
    \item \textbf{2-hop:} SDM-InstructGLM reaches 84.92\%, an improvement of \textbf{16.3\%}.  
    \item \textbf{3-hop:} SDM-InstructGLM maintains 83.81\%, \textbf{13.1\%} higher than InstructGLM.  
    \item The substantial improvements suggest that SDM-InstructGLM effectively incorporates higher-order structural dependencies and node interactions, making it superior in long-range information aggregation. By leveraging feature similarity and degree importance, our mechanism effectively enhances class differentiation in CORA. Given that CORA has seven classes and a relatively sparse graph structure, these weighting strategies improve node separability, address class boundary challenges\cite{luan2024}, and ultimately enhance classification performance.
\end{itemize}

\noindent\textbf{\textsc{PubMed} Dataset:} 
\begin{itemize}
    \item \textbf{1-hop:} SDM-InstructGLM achieves 91.70\%, slightly surpassing InstructGLM by \textbf{0.15\%}.  
    \item \textbf{2-hop:} SDM-InstructGLM reaches 91.48\%, marking a \textbf{0.67\%} increase.  
    \item \textbf{3-hop:} SDM-InstructGLM attains 91.23\%, improving by \textbf{0.38\%}.  
    \item The gains on \textsc{PubMed} are less pronounced compared to \textsc{Cora}, likely due to \textsc{PubMed} having a fewer number of node classes, which makes classification tasks less difficult by nature due to higher baseline accuracy and the strong semantic signals in features. This causes less overlap in decision boundaries, as previously pointed out in \cite{luan2024}. As the diameters of the two datasets do not differ significantly, it does not seem to affect our outcome. Nevertheless, SDM-InstructGLM maintains a consistent edge, demonstrating more stable performance across varying hop distances.
\end{itemize}

\subsubsection{Interpretation and Key Findings}
\begin{itemize}
    \item \textbf{Significant performance boost on \textsc{Cora}:} The large improvements suggest that SDM-InstructGLM excels at capturing structural dependencies in sparser graphs like \textsc{Cora}, where node connectivity plays a crucial role in classification accuracy.  
    \item \textbf{Consistent gains on \textsc{PubMed}:} While the performance gap is smaller on \textsc{PubMed}, SDM-InstructGLM still achieves better generalization, indicating that its structured node ordering and hop-aware selection effectively retain semantic relevance without excessive noise.  
    \item \textbf{Better multi-hop reasoning:} The model’s robustness across multiple hop distances reinforces that SDM-InstructGLM can balance local (1-hop) and global (3-hop) structural awareness, a key advantage over conventional LLM-based graph representations.  
\end{itemize}

These results validate the effectiveness of our biased random walk, structured node ordering, and hop-aware node selection strategies in improving graph-based LLM representations. The improvements observed in SDM-InstructGLM suggest that it is a more scalable and adaptive solution for graph representation learning.

\subsection{Ablation Study (RQ2)}

To evaluate the impact of different components in SDM-InstructGLM, we conduct ablation studies on similarity-based and degree-based weighting in the biased random walk mechanism, as well as analyze performance under limited prompt node conditions.

\subsubsection{Ablation Study on Similarity Scores}
Table~\ref{ablation1} investigates the effect of disabling cosine similarity and node degree on model accuracy for the \textsc{Cora} and \textsc{PubMed} datasets.

\textbf{Findings:}
\begin{itemize}
    \item \textbf{Full Model (None Disabled):} Achieves the highest accuracy 83.49\% on \textsc{Cora}, 91.47\% on \textsc{PubMed}), confirming that both similarity-based transitions and degree-based weighting contribute significantly.
    \item \textbf{Cosine Similarity Disabled:} Leads to a \textbf{20.35\%} drop on \textsc{Cora}, but has no significant effect on \textsc{PubMed}. This suggests that \textsc{Cora} relies heavily on feature similarity, while \textsc{PubMed}’s rich textual information\cite{pubmed} reduces dependence on similarity weighting.
    \item \textbf{Node Degree Disabled:} Causes a similar \textbf{20.57\%} accuracy drop on \textsc{Cora}, indicating that degree-based importance plays a major role in sparse graphs. On \textsc{PubMed}, accuracy declines slightly (\textbf{0.53\%}), suggesting structural connectivity is helpful but less critical.
\end{itemize}

These results indicate that both cosine similarity and node degree are essential in sparser datasets like \textsc{Cora}, while \textsc{PubMed} benefits less due to its stronger textual features.

\subsubsection{Ablation Study with Limited Prompt Node Numbers}
Table~\ref{ablation2} examines how SDM-InstructGLM performs with a limited number of prompt nodes in the \textsc{PubMed} dataset, with the limits of 5, 25, and 125 structures for 1-hop, 2-hop, and 3-hop structures respectively. Accuracy is compared across 1-hop, 2-hop, and 3-hop structures against the original InstructGLM.

\textbf{Findings:}
\begin{itemize}
    \item \textbf{SDM-InstructGLM consistently outperforms InstructGLM across all hop distances.}
    \item The highest gain is in the 1-hop setting (\textbf{+2.18\%}), showing that SDM-InstructGLM extracts more meaningful information from immediate neighbors.
    \item Multi-hop performance stabilizes, with SDM-InstructGLM maintaining a \textbf{+0.97\%} improvement at 3-hop, suggesting efficient long-range dependency learning.
    \item The results confirm that SDM-InstructGLM learns effectively even with fewer nodes, demonstrating its robustness under token constraints.
\end{itemize}

These ablation studies validate the effectiveness of biased random walk with similarity and degree weighting and structured node selection in SDM-InstructGLM. Key takeaways include:
\begin{itemize}
    \item Both similarity-based and degree-based biasing significantly improve performance, particularly in sparse graphs like \textsc{Cora}.
    \item SDM-InstructGLM outperforms InstructGLM even when fewer prompt nodes are available, demonstrating better generalization and efficiency.
    \item The model effectively balances local and global context, achieving consistent improvements in multi-hop reasoning.
\end{itemize}

\newcolumntype{Z}{>{\centering}X} 
\newcolumntype{Y}{>{\centering\arraybackslash}X} 
\begin{table}[t]
	\centering
	\caption{Ablation Study Results on Similarity Scores}
	\renewcommand{\arraystretch}{1.5} 
	\begin{tabularx}{0.75\columnwidth}{@{}Z| *2{Y}@{}}
		\hline 
		\hline	
		\multirow{2}*{\shortstack{Methodology}} & \multicolumn{2}{c}{Accuracy (\%) by Dataset}\\ 
        \cline{2-3}
        ~ & \textsc{Cora} & \textsc{PubMed} \\
		\hline		
		\textbf{SDM-InstructGLM} & \textbf{83.49} & 91.47 \\ 
        w/o Cos Similarity & 63.14 & \textbf{91.48} \\
        w/o Node Degree & 62.92 & 90.94 \\
		\hline
	\end{tabularx}
    \label{ablation1}
\end{table}

\begin{table}[t]
	\centering
        \captionsetup{justification=centering}
	\caption{Ablation Study with Limited Prompt Node Numbers (PubMed)}
	\renewcommand\arraystretch{1.3}
	\begin{tabular}{c|ccc}
		\hline 
		\hline	
		\multirow{2}*{Model} & \multicolumn{3}{c}{Accuracy(\%) by \# Hops}\\ 
        \cline{2-4}
        ~ & 1 & 2 & 3\\
		\hline		
		InstructGLM (Original)  &  88.26 &  89.53 & 89.55 \\
		SDM-InstructGLM   & \textbf{90.44}  & \textbf{90.72} & \textbf{90.52} \\
		\hline
	\end{tabular}
		
	\label{ablation2}
\end{table}

\section{Conclusion}

SDM-InstructGLM enhances graph representation learning by integrating biased random walks, structured node ordering, and hop-aware selection, leading to improvements over InstructGLM. Experiments on \textsc{Cora} and \textsc{PubMed} confirm its effectiveness in preserving structural and semantic information while maintaining robustness under token constraints. Ablation studies highlight the importance of cosine similarity and degree-based weighting, particularly in sparse graphs. These results demonstrate that SDM-InstructGLM is well-suited for capturing both local and global dependencies, making it a strong candidate for improving graph-based learning tasks.

Future work will focus on extending SDM-InstructGLM to larger and more complex graphs, exploring its adaptability across different domains such as social networks, knowledge graphs, and biological networks. Additionally, investigating more efficient training strategies and refining prompt-based learning approaches will help enhance scalability and computational efficiency.

Once computational constraints are alleviated, we plan to address our work's limitations, in identifying the specific causes of our result. We also plan to conduct experiments on a broader range of benchmark datasets to further evaluate the generalization ability of SDM-InstructGLM across diverse graph structures. By addressing these challenges, SDM-InstructGLM can further solidify its role as a versatile and effective foundation model for graph-based machine learning. Moreover, through additional experiments, we will be able to identify the specific characteristics within the dataset that brings advantage to our node-ordering method over the random-walk method.

\section*{Acknowledgment}

This research was supported by the Department of Computer Science at Trinity College, Hartford, Connecticut. The authors would like to extend their special thanks to Dr. Madalene Spezialetti, Chair of the Computer Science Department, for her generous support throughout the course of this work.

\bibliographystyle{IEEEtran}
\bibliography{main}

\end{document}